\documentclass[10pt,twocolumn,letterpaper]{article}

\usepackage{cvpr}              
\definecolor{cvprblue}{rgb}{0.21,0.49,0.74}
\usepackage[pagebackref,breaklinks,colorlinks,allcolors=cvprblue]{hyperref}

\def\paperID{44381} 
\def\confName{CVPR}
\def\confYear{2026}

\title{\method{}: Towards Medical Vision-Language Model's Clinical Intelligence}

\author{
Woohyeon Park$^{1}$ \quad
Jaeik Kim$^{1}$ \quad
Sunghwan Steve Cho$^{1}$ \quad
Pa Hong$^{2}$ \quad
Wookyoung Jeong$^{3}$ \\
Yoojin Nam$^{2}$ \quad
Namjoon Kim$^{2}$ \quad
Ginny Y. Wong$^{4}$ \quad
Ka Chun Cheung$^{4}$ \quad
Jaeyoung Do$^{1}$ \\
$^{1}$AIDAS Laboratory, Seoul National University \quad
$^{2}$Samsung Changwon Hospital \\
$^{3}$Samsung Medical Center \quad
$^{4}$NVIDIA, Santa Clara, USA \\
{\tt\small \{woohyeon, jake630, steve97, knj01, jaeyoung.do\}@snu.ac.kr} \quad 
{\tt\small jeongwk@gmail.com} \quad \\
{\tt\small \{papa.hong, yoojin8998.nam\}@samsung.com} \quad
{\tt\small \{gwong, chcheung\}@nvidia.com} \\
\url{https://github.com/AIDASLab/Medic-AD}
}

\usepackage{graphicx}
\usepackage{amssymb}
\usepackage{amsmath}
\usepackage{dsfont}
\usepackage{multirow}
\usepackage{float}
\usepackage{lipsum}
\usepackage{nicefrac}
\usepackage{circledsteps}
\usepackage{enumitem}
\usepackage{colortbl}
\usepackage{graphicx}
\usepackage{booktabs}
\usepackage{siunitx}
\usepackage{breqn}
\usepackage{lipsum}
\usepackage{adjustbox}
\usepackage{duckuments}
\usepackage{tikzducks}
\usepackage{tablefootnote}
\usepackage{pifont} 
\usepackage{makecell}
\usepackage{tikz}
\usepackage{algorithm}
\usepackage{algorithmic}
\usepackage{tcolorbox}
\usepackage{dsfont}

\definecolor{lightgray}{gray}{0.95}
\definecolor{midgray}{gray}{0.55}
\definecolor{steelblue}{HTML}{4D82B7}
\definecolor{davysgrey}{rgb}{0.33, 0.33, 0.33}
\definecolor{LightCyan}{rgb}{0.88,1,1}
\definecolor{ao(english)}{rgb}{0.0, 0.5, 0.0}

\usepackage[first=0, last=9]{lcg}

\newcommand{\Star}[1]{#1\ensuremath{^*}\kern-\scriptspace}

\makeatletter
\DeclareRobustCommand\onedot{\futurelet\@let@token\@onedot}
\def\@onedot{\ifx\@let@token.\else.\null\fi\xspace}

\definecolor{algc1}{HTML}{f7d779}
\definecolor{algc2}{HTML}{9fc5fc}

\renewcommand{\algorithmiccomment}[1]{\bgroup\hfill $\triangleright$ ~#1\egroup}

\usepackage{array}
\newcommand{\PreserveBackslash}[1]{\let\temp=\\#1\let\\=\temp}
\newcolumntype{C}[1]{>{\PreserveBackslash\centering}p{#1}}
\newcolumntype{R}[1]{>{\PreserveBackslash\raggedleft}p{#1}}
\newcolumntype{L}[1]{>{\PreserveBackslash\raggedright}p{#1}}

\newcommand{\method}{\textsc{Medic-AD}\xspace}
\newcommand{\anotok}{\texttt{<Ano>}\xspace}
\newcommand{\difftok}{\texttt{<Diff>}\xspace}

\usepackage[normalem]{ulem}

\begin{document}
\maketitle
\begin{abstract}
Lesion detection, symptom tracking, and visual explainability are central to real-world medical image analysis, yet current medical Vision-Language Models (VLMs) still lack mechanisms that translate their broad knowledge into clinically actionable outputs. To bridge this gap, we present \method{}, a clinically oriented VLM that strengthens these three capabilities through a stage-wise framework. First, learnable anomaly-aware tokens (\anotok{}) encourage the model to focus on abnormal regions and build more discriminative lesion centered representations. Second, inter-image difference tokens (\difftok{}) explicitly encode temporal changes between studies, allowing the model to distinguish worsening, improvement, and stability in disease burden. Finally, a dedicated explainability stage trains the model to generate heatmaps that highlight lesion-related regions, offering clear visual evidence that is consistent with the model's reasoning. Through our staged design, \method{} steadily boosts performance across anomaly detection, symptom tracking, and anomaly segmentation, achieving state-of-the-art results compared with both closed source and medical-specialized baselines. Evaluations on real longitudinal clinical data collected from real hospital workflows further show that \method{} delivers stable predictions and clinically faithful explanations in practical patient-monitoring and decision-support workflows.
\end{abstract}
\begin{figure}[t]
    \centering
    \includegraphics[width=0.5\textwidth]{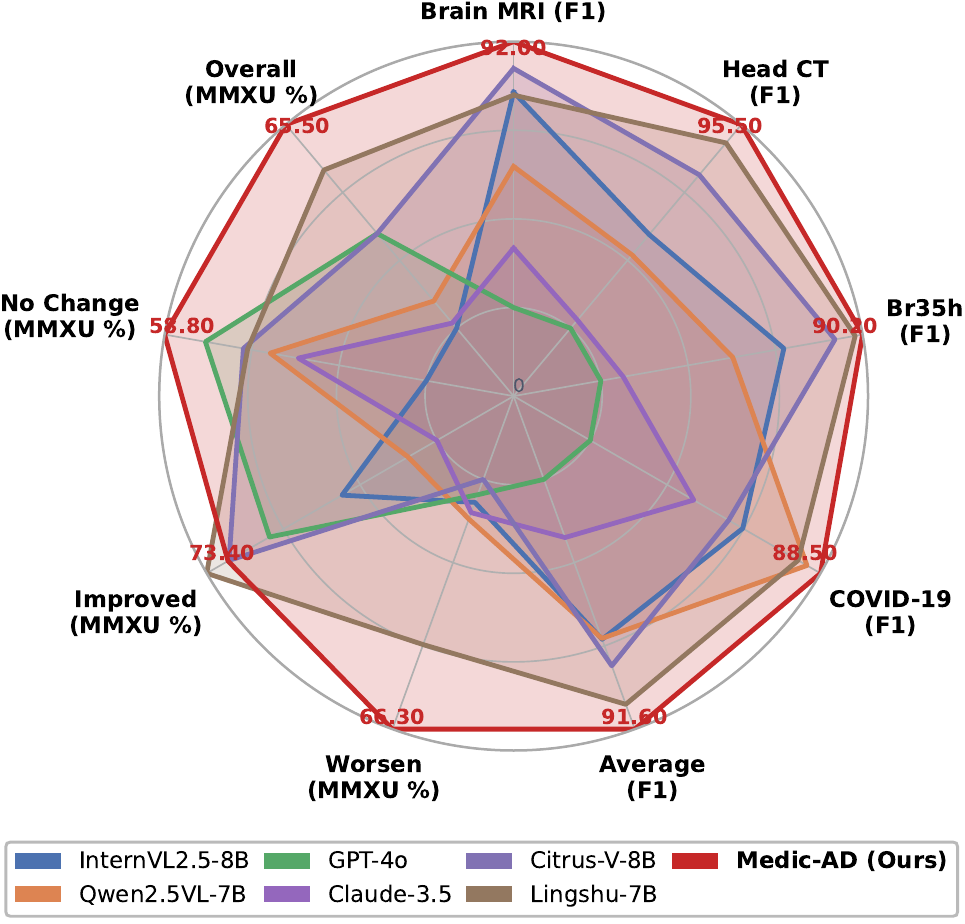}
    \caption{Overall performance of VLMs on Medical Anomaly Detection and Medical Symptom Tracking (MMXU~\cite{mmxu}).}
    \label{fig1}
    \vspace{-0.5em}
\end{figure}
\label{subsec:architecture}

\section{Introduction}
\label{sec:1.introduction}
Vision–Language Models (VLMs) have rapidly evolved~\cite{llava, minigpt, instructblip} from simple image–text tasks such as visual question answering (VQA)~\cite{vqa} and captioning~\cite{captioning} to more advanced capabilities including visual grounding~\cite{visual_grounding, glamm} and multi–image reasoning~\cite{llava_ov, qwen2.5vl, internvl2.5}. These advances have inspired the emergence of \textit{Medical Foundation VLMs}~\cite{llava_med, xraygpt, med_flamingo}, which aim to integrate visual and textual medical knowledge for comprehensive diagnostic understanding. Trained on large-scale image-report pairs 
and multimodal instructions, these models have achieved strong results across tasks such as disease classification, report generation, and Med–VQA, demonstrating the promise of language-driven clinical reasoning.

However, most Medical Foundation VLMs remain optimized for broad medical knowledge coverage rather than real clinical application~\cite{lingshu, citrus_v}. Their training typically relies on long–form captioning, OCR–based instruction tuning, and medical chain–of–thought reasoning that enhance generic reasoning ability, but overlook key properties required for real-world clinical workflows~\cite{radiology}: (1) accurate lesion detection, (2) reliable temporal symptom tracking, and (3) transparent visual explainability of the reasoning process. Addressing these limitations demands a paradigm shift from generalized intelligence toward clinically grounded perception, and understanding.

To that end, we explore three research questions guiding the design of a clinically usable medical VLM.

\vspace{-1.0em}
\paragraph{RQ1: How can lesion and symptom recognition be improved in VLMs for real clinical settings?}
Even as medical VLMs expand in knowledge, accurate abnormality detection remains essential for safe deployment. We define an \emph{abnormality} as any pathological deviation within an image and propose to enhance this recognition through explicit anomaly–aware representations. By injecting learnable \anotok{} tokens into the transformer layers, the model highlights abnormal regions and strengthens its discriminative reasoning. Experiments on brain MRI, head CT, and chest X-ray datasets show that this design achieves strong performance in medical anomaly detection.

\vspace{-1.0em}
\paragraph{RQ2: How can a VLM disentangle temporal medical images to enable more accurate symptom tracking?}
Existing foundation models that support multi–image inputs typically concatenate visual features, thus failing to capture the temporal progression between scans. To model clinically meaningful changes, we introduce \difftok{} tokens that compare anomaly features extracted from multiple images of the same patient. These representations allow the model to reason about whether a condition has worsened, improved, or remained unchanged. On benchmarks such as MMXU~\cite{mmxu}, which assess longitudinal symptom understanding, our approach achieves superior performance, highlighting its effectiveness as a practical clinical tool for patient monitoring.

\vspace{-1.0em}
\paragraph{RQ3: How can visual explainability be integrated into medical VLM reasoning?}
Explainability is indispensable for clinical decision-making. To visually justify model's predictions, we design a heatmap decoder that fuses anomaly features and visual features to 
generate visualization maps highlighting regions responsible for each prediction. These region-level explanations enhance transparency by providing visual evidence for both lesion detection and change assessment, ultimately bridging the gap between black-box reasoning and clinical trust.\\

These three research directions culminate in \textbf{\method{}}, a stage-wise medical VLM with clinical intelligence, designed to integrate anomaly detection, temporal reasoning, and visual explainability. \method{} is trained in three stages:
\textit{Stage 1: Anomaly Detection.} Learn discriminative abnormality embeddings via injected anomaly tokens, \anotok{}, adapting contrastive architecture between normal and abnormal regions for enhancing sensitivity to pathological cues.
\textit{Stage 2: Difference Reasoning.} Encode cross-scan variations using \difftok{} tokens that disentangle temporal progression of abnormal features and enable fine-grained symptom tracking. \textit{Stage 3: Visual Explainability.} Generate visual evidence heatmaps that ground textual outputs on abnormal regions, ensuring verifiable reasoning.

Through extensive evaluation, \method{} consistently demonstrates state-of-the-art (SOTA) performance across diverse medical modalities and tasks as shown in Fig.~\ref{fig1}. It outperforms medical foundation models~\cite{llava_med, lingshu, citrus_v}, anomaly–specialized models~\cite{anomaly_gpt, anomalyov}, and closed-source counterparts~\cite{gpt4o, claude} in both lesion detection, and temporal symptom reasoning. Moreover, \method{} delivers superior visual explainablity generating spatially grounded explanations that align model decisions with clinical evidence. Beyond numerical gains, its stage–wise design 
encodes the clinical diagnostic workflow—detect, compare, explain—into the model’s learning curriculum, transforming general-purpose vision–language understanding into clinically actionable intelligence.

Our main contributions are as follows:

\begin{itemize}
    \item We present a unified, stage-wise framework ({\method{}}) that integrates anomaly detection, longitudinal reasoning, and visual grounding to enable explainable medical inference.
    \item We introduce anomaly- and difference-token mechanisms that endow medical VLMs with explicit lesion sensitivity and temporal reasoning capability.
    \item We conduct comprehensive evaluations on multiple medical tasks, as well as on real longitudinal datasets from hospital sites, demonstrating superior reliability and usability compared to both open- and closed-source foundation models, and showcasing deployment readiness for real-world clinical practice workflows. 
\end{itemize}

\section{Related Works}
\label{sec:2.related_works}

\subsection{Vision-Language Models for Medical Imaging}
\label{subsec:2.1}
Vision Language Models (VLMs)~\cite{llava, minigpt, instructblip, llava_ov, qwen2.5vl, internvl2.5} have unified visual and textual reasoning across domains through large-scale contrastive or instruction-tuned learning. Building upon these, medical VLMs have adapted multimodal alignment to clinical imaging and reporting tasks.
Early medical VLMs focused on contrastive alignment and report-level representation learning~\cite{medclip, bio_vil}, while later instruction-tuned architectures expanded multimodal reasoning through larg scale medical--text pretraining~\cite{llava_med, medtrinity, medgemma, slake}.  
More recently, Lingshu and Citrus-V~\cite{lingshu, citrus_v}, built on Qwen-VL 2.5~\cite{qwen2.5vl}, introduced multi-stage training with shallow/deep alignment, medical instruction tuning, and reinforcement learning with verifiable rewards, achieving state-of-the-art results on single-image medical VQA benchmarks such as SLAKE, PathVQA, VQA-RAD, and OmniMedVQA~\cite{slake, pathvqa, vqa_rad, omnimedvqa}. These datasets collectively evaluate anatomical localization, factual consistency, and report generation, forming the empirical foundation for single-image medical VLMs.
Beyond single-image understanding, medical VLMs have advanced toward difference-aware reasoning, modeling longitudinal changes and disease progression between paired studies. Generic difference captioning frameworks describe visual changes across image pairs~\cite{mccformers, idcpcl}, whereas medical-specific longitudinal reasoning integrates anatomical or report-based temporal modeling~\cite{ekaid, plural}.  
A recent unified framework~\cite{reportgen2025} proposes a \textit{Report Generator--Answer Generator (RG--AG)} architecture. While these studies have achieved meaningful progress in understanding longitudinal medical images, they remain largely task-specific, focusing on objectives such as VQA and report generation, without fully leveraging the extensive medical knowledge and generative reasoning capabilities offered by Medical Foundation VLMs.

\subsection{Zero-Shot Anomaly Detection}
\label{subsec:2.2}
Traditional anomaly detection (AD) methods primarily focused on low-level visual irregularities in industrial datasets such as MVTec-AD~\cite{mvtc_ad} and VisA~\cite{spot}, later extending toward zero-shot recognition through text--image alignment. CLIP-based approaches introduced adapter- or prompt-based mechanisms for open-vocabulary detection~\cite{anomalyclip, adaclip, mvda}, while Q-Former-based~\cite{blip2} architectures further connected anomaly detection and instruction tuning~\cite{anomalyov, anomaly_gpt} in general AD tasks. In the medical context, unified benchmarks such as BMAD~\cite{bmad} integrate diverse medical anomaly detection datasets—covering Brain MRI, Chest X-Ray, Liver CT, Retinal OCT, and Pathology—into a single evaluation framework. BMAD consolidates various modality-specific datasets to enable consistent cross-dataset and cross-organ evaluation under a unified protocol. In parallel, chest-specific datasets such as ChestX-Det~\cite{chestxdet} further provide detailed pixel-level annotations for thoracic disease localization and anomaly segmentation. Collectively, these works mark a shift from handcrafted features to medically grounded anomaly understanding.

\subsection{Explainability in Vision--Language Models}
\label{subsec:2.3}
Explainability has become a key criterion for assessing the reliability of VLMs, especially in safety-critical domains such as medicine.  
Recent VLMs utilize cross-attention heatmaps and token-conditioned activations to visualize how linguistic tokens attend to visual regions during reasoning, thereby revealing the internal correspondence between textual semantics and spatial evidence~\cite{interpretingCLIP, llava, alignBeforeFuse, kosmosG}.  
Such mechanisms improve transparency and enable systematic model auditing and error analysis by linking visual attention patterns with generated textual outputs.

In medical imaging, explainability has been advanced through explicit grounding and concept-level alignment.  
Grounded VLMs explicitly associate textual rationales with anatomical regions~\cite{bio_vil, radGraph, reportgen2025}, while concept-disentanglement approaches align clinical entities with visual concepts to enhance explainability and trustworthiness~\cite{deal}. Building upon these developments, we extend visual grounding beyond a mere auxiliary visualization tool. In our medical VLM, explainability is achieved by grounding the reasoning process on anomalous features such as lesions or symptoms, thereby providing explicit visual evidence that supports the model’s clinical conclusions and ensuring clinically verifiable visual explainability.

\begin{figure}[t]
    \centering
    \includegraphics[width=0.47\textwidth]{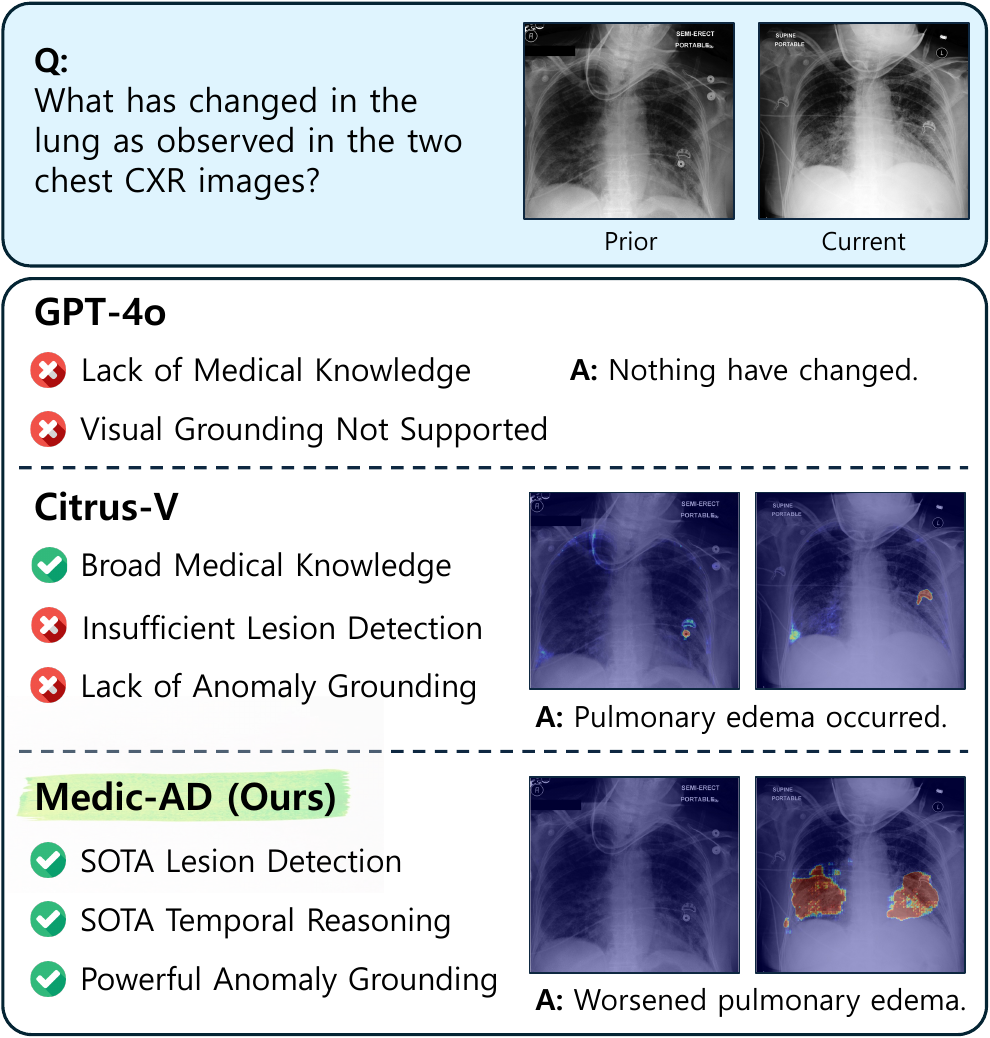}
    \vspace{-0.5em}
    \caption{Comparison of VLMs on clinical applications. Medic-AD provides stronger lesion detection, temporal reasoning, and visual grounding than GPT-4o~\cite{gpt4o} and Citrus-V~\cite{citrus_v}.}
    \label{fig2}
    \vspace{-1.0em}
\end{figure}
\section{Methodology}
\label{sec:3.methodology}

\subsection{Overview}
\label{subsec:3.1.overview}
Standard VLMs encode an input image $\mathbf{I}$ and textual instruction $\mathbf{T}$ into a joint multimodal sequence to generate a response $\mathbf{R}$:
\begin{align}
\mathbf{R} = f_l\big( [\, f_p(\mathbf{V}) \,;\, \text{Emb}(\mathbf{T}) \,] \big),
\end{align}
\begin{figure*}[t]
    \centering
    \includegraphics[width=1.0\linewidth]{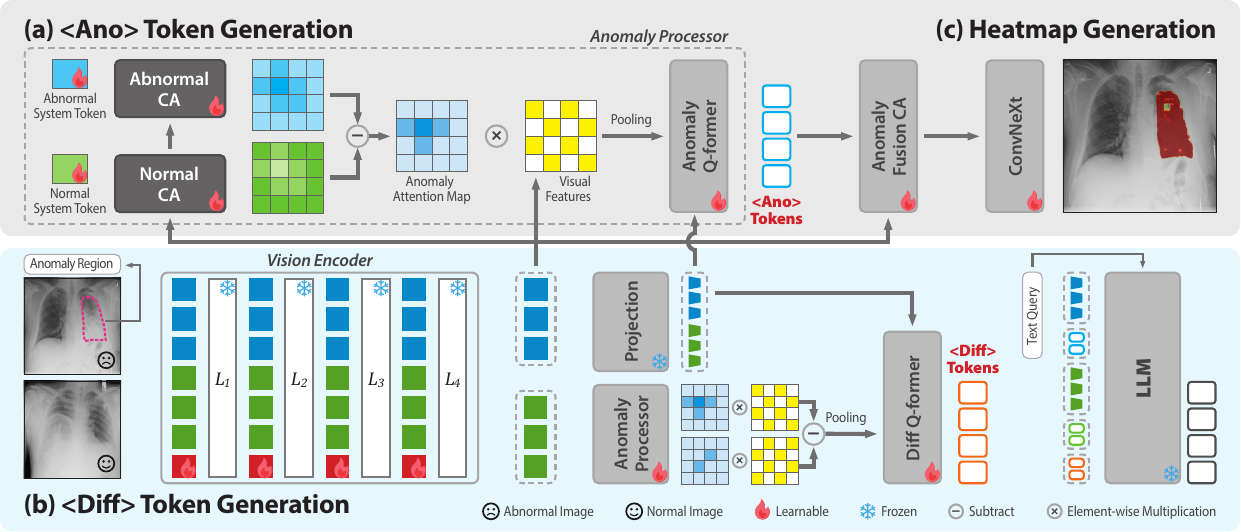}
    \caption{
    Architecture of \method{}. 
    (a) Stage~1: \anotok{} Token Generation,
    (b) Stage~2: \difftok{} Token Generation, and
    (c) Stage~3: Heatmap Generation illustrate each stage of the proposed framework.
    Note that CA denotes Cross-Attention.}
    \vspace{-1.0em}
    \label{fig3}
\end{figure*}
where $f_p(\cdot)$ denotes a visual projection layer that maps visual features $\mathbf{V}$, extracted from $\mathbf{I}$ via a vision encoder, into the text embedding space, and $f_l(\cdot)$ represents the large language model (LLM). While this general formulation supports broad multimodal reasoning, it lacks the inductive biases required for clinically meaningful perception including reliable lesion localization, temporal tracking, and visual justification, all of which are essential for trustworthy decision support.

\method{} extends this paradigm into a clinically grounded reasoning framework through a stage-wise optimization pipeline, composed of three progressive stages where each stage incrementally enhances the model’s reasoning capability. Stage 1 learns anomaly-aware representations that encode lesion-specific semantics. Stage 2 builds on these representations to disentangle temporal variations between prior and current studies, yielding difference tokens. Stage 3 introduces grounding supervision that aligns the learned anomaly features with spatial heatmaps, enabling visually verifiable predictions.

We first introduce anomaly-aware tokens, $\anotok{}$, derived from a cross-attention mechanism applied to the visually enhanced feature representation $\mathbf{V^*}$. 
Here, $\mathbf{V^*}$ denotes the anomaly-augmented visual features obtained by incorporating \textit{visual soft prompts}~\cite{vpt} into the original visual embeddings $\mathbf{V}$. 
This process encourages the model to emphasize lesion-relevant regions and discriminative cues. 
The resulting tokens are concatenated with the visual and textual embeddings as
\vspace{-0.2em}
\begin{align}
\mathbf{R} = f_l\big( [\, f_p(\mathbf{V^*}) \,;\, \anotok{} \,;\, \text{Emb}(\mathbf{T}) \,] \big).
\end{align}

\vspace{-0.2em}
Next, to model temporal changes between prior and current images, our model learns anomaly-aware representations across time and produces \difftok{} tokens.
Given two input images, their corresponding anomaly-augmented visual features, $\mathbf{V^*_1}$ and $\mathbf{V^*_2}$, are extracted by the vision encoder and formulated as
\begin{align}
\begin{split}
\mathbf{R} = f_l\big( [\, &f_p(\mathbf{V^*_1}) \,;\, \anotok{} \,;\, f_p(\mathbf{V^*_2}) \,;\, \anotok{} \,; \\
&\text{Emb}(\mathbf{T}) \,; \, \difftok{} \,] \big),
\end{split}
\end{align}
following the modified chat template as illustrated in Appendix Sec.~\ref{sec:A.chat}.
Finally, the anomaly-aware tokens, \anotok{}, are fed into a heatmap decoder $f_h(\cdot)$ together with the corresponding visual features  $\mathbf{V^*}$ to generate grounding maps $\mathbf{M}$. These maps provide region-level visual evidence that supports and justifies textual predictions as
\begin{align}
\mathbf{M} = f_h\big( [\, f_p(\mathbf{V^*}) \, ;\, \anotok{} \,] \big).
\end{align}

\subsection{Architecture and Stage-wise Training}
\label{subsec:3.2.training_pipelines}
In this section, we present the detailed architecture of \method{} and the corresponding training pipelines designed to implement the framework illustrated in Sec.~\ref{subsec:3.1.overview}. 
Each stage progressively enhances the capability of the baseline Medical Foundation VLM, Lingshu~\cite{lingshu}, which is a strong backbone model pretrained on large-scale medical data, by introducing specialized modules for anomaly reasoning, temporal difference analysis, and visual explainability.
\paragraph{Stage 1: Anomaly Detection.}
The first stage focuses on training an \textit{Anomaly Processor} that produces \anotok{} tokens, compact latent representations capturing lesion-related semantics, as illustrated in Fig.~\ref{fig3}~(a). These tokens are constructed through two learnable system tokens, the \textit{Abnormal System Token} and \textit{Normal System Token}. They interact with multi-scale visual features extracted from four intermediate layers of the vision encoder via a cross-attention mechanism, producing \textit{Abnormal} and \textit{Normal Attention Scores} for each visual patch.

To preserve the pretrained vision encoder’s representational stability while adapting it for anomaly detection, we adopt \textit{Visual Soft Prompt Tuning}~\cite{vpt} to the selected four layers instead of fully updating their parameters. Unlike conventional attention head using Softmax normalization, our design applies Sigmoid activation to obtain patch-wise anomaly probabilities.
The difference between abnormal and normal attention weights yields an \textit{Anomaly Attention Map}, which reflects the likelihood of each patch being abnormal. This map modulates the original visual features through element-wise multiplication, adjusting their magnitudes according to anomaly salience.

Subsequently, 2D global pooling is performed over the modulated visual features to derive \textit{Anomaly Queries}. 
These queries are passed through an \textit{Anomaly Q-Former}, where the LLM-projected visual tokens act as keys and values. 
The Anomaly Q-Former outputs are then fed into a 2-layer MLP to yield the final Anomaly-aware tokens, \anotok{}, which are used in both downstream LLM inference and later heatmap generation in Stage~3.

Training for this stage utilizes a diverse collection of medical anomaly datasets spanning MRI, X-ray, and CT modalities, including \textbf{BMAD}, \textbf{ChestX-Det}~\cite{bmad, chestxdet}, as well as multimodal VQA datasets such as \textbf{SLAKE}, \textbf{PathVQA}, and \textbf{VQA-RAD}~\cite{slake, pathvqa, vqa_rad}, ensuring both robust visual grounding and generalizable medical reasoning capability.
\paragraph{Stage 2: Difference Reasoning.}
The second stage focuses on modeling inter-image differences to analyze disease progression over time. Here, the goal is to learn difference tokens, \difftok{}, that \textit{disentangle} the variations in abnormal regions across time or paired studies (e.g., follow-up vs. baseline scans), effectively separating genuine pathological progression from visual or acquisition-related noise.

As shown in Fig.~\ref{fig3}~(b), the modulated visual features derived in Stage~1 from two images are contrasted and disentangled through a \textit{Diff Q-Former}, which isolates lesion-specific change patterns. 
Then, each image's projected visual tokens, $f_p(\mathbf{V^*_1})$ and $f_p(\mathbf{V^*_2})$, serves as keys and values to encode structured inter-image relationships. 
Passing these through the Diff Q-Former and a subsequent 2-layer MLP yields the difference tokens, \difftok{}, which are appended to the multimodal input sequence. By explicitly isolating temporal anomalies from static visual context, this stage enables the LLM to reason over fine-grained temporal variations in lesion appearance or intensity, thereby enhancing its ability for longitudinal disease reasoning.

Stage~2 training requires temporally paired or longitudinal datasets. We use \textbf{MIMIC-Diff-VQA}\cite{ekaid}, a dataset built for multi-image reasoning in clinical follow-up scenarios, allowing the model to learn spatial correspondence and temporal progression patterns in real patient studies.
\paragraph{Stage 3: Visual Explainability.}
The final stage introduces a heatmap generation module designed to achieve visual grounding and enhance explainability. While prior Medical Foundation VLMs often rely on pretrained vision decoders (e.g., SAM2~\cite{sam2} used in Citrus-V~\cite{citrus_v}), our approach leverages the learned \anotok{} tokens and ConvNeXt-based \cite{convnext} segmentation head to directly link visual reasoning with evidence.

As illustrated in Fig.~\ref{fig3}~(c), we fuse \anotok{} tokens with the vision encoder’s intermediate feature maps via a fusion block, reinforcing the model’s focus on lesion-relevant regions driving the LLM’s prediction.
The fused features are then processed by a compact ConvNeXt-based segmentation head to generate a heatmap $\mathbf{M}$ spatially aligned with the input image. 
This heatmap is overlaid on the original image to provide region-level visual evidence that supports the textual output, thereby connecting model reasoning with clinically observable cues.

Stage 3 is trained on datasets with pixel-level segmentation masks, such as selected subsets of \textbf{BMAD} and \textbf{ChestX-Det}~\cite{bmad, chestxdet}. Leveraging anomaly-token-guided fusion, our model achieves substantially improved anomaly-localization accuracy compared with recent grounding-based medical VLMs, as demonstrated in Sec.~\ref{subsec:4.3.seg}. For the more detailed training configurations of each stages, please see Appendix Sec.~\ref{sec:B.training}.

\section{Experiments}
\label{sec:4.experiments}

\newcolumntype{F}{>{\columncolor{gray!10}}c}

\renewcommand{\arraystretch}{0.85}
\begin{table*}[t]
\small
\centering
\caption{Results on Zero-shot Medical Anomaly Detection (Brain MRI~\cite{brainmri}, Head CT~\cite{headct}, Br35h~\cite{Br35h}, and COVID-19~\cite{covid19}). For each dataset, we report Precision, Recall, and F1. The rightmost column shows the average F1 over all tasks.}
\label{tab:tab1}
\setlength{\tabcolsep}{0.55em}
\begin{tabular}{l|l|c| cc|F| cc|F| cc|F| cc|F| c}
\toprule
\multirow{2}{*}{\textbf{Category}} & \multirow{2}{*}{\textbf{Model}} & \multirow{2}{*}{\textbf{Size}} &
\multicolumn{3}{c}{\textbf{Brain MRI}} &
\multicolumn{3}{c}{\textbf{Head CT}} &
\multicolumn{3}{c}{\textbf{Br35h}} &
\multicolumn{3}{c}{\textbf{COVID-19}} &
\multirow{2}{*}{\textbf{Avg. F1}} \\
\cmidrule(lr){4-6}\cmidrule(lr){7-9}\cmidrule(lr){10-12}\cmidrule(lr){13-15}
 &  &  & \textbf{Prec.} & \textbf{Rec.} & \textbf{F1}
 & \text{Prec.} & \text{Rec.} & \textbf{F1}
 & \text{Prec.} & \text{Rec.} & \textbf{F1}
 & \text{Prec.} & \text{Rec.} & \textbf{F1} &  \\
\midrule

\multirow{2}{*}{\textbf{General}}
& Qwen2.5-VL      & 7B  & 76.1 & 92.9 & 83.6 & 61.6 & 100.0 & 76.3 & 66.4 & 94.5 & 78.0 & 88.6 & 83.4 & 85.9 & 81.0 \\
& InternVL2.5    & 8B  & 81.2 & 97.4 & 88.6 & 67.6 & 96.0 & 79.3 & 71.2 & 99.0 & 82.8 & 94.8 & 60.2 & 73.6 & 81.1 \\
\midrule

\multirow{2}{*}{\textbf{Closed}}
& GPT-4o         & --  & 59.3 & 98.6 & 74.1 & 48.7 & 100.0 & 65.5 & 48.9 & 99.9 & 65.6 & 29.1 & 92.9 & 44.4 & 62.4 \\
& Claude-3.5     & --  & 64.0 & 100.0 & 78.1 & 50.0 & 100.0 & 66.7 & 51.2 & 100.0 & 67.7 & 47.2 & 100.0 & 64.2 & 69.2 \\
\midrule

\multirow{2}{*}{\textbf{Anomaly}}
& AnomalyGPT     & 13B & 61.0 & 96.8 & 74.8 & 50.8 & 97.0 & 66.7 & 47.2 & 88.6 & 61.6 & 33.9 & 57.2 & 42.5 & 61.4 \\
& Anomaly-OV     & 7B  & 53.1 & 71.6 & 61.0 & 39.8 & 66.0 & 49.6 & 42.2 & 73.0 & 53.5 & 29.6 & 47.0 & 36.3 & 50.1 \\
\midrule

\multirow{4}{*}{\textbf{Medical}}
& LLaVA-MED      & 7B  & 69.0 & 95.5 & 80.1 & 54.1 & 77.9 & 63.8 & 60.8 & 92.7 & 73.4 & 46.1 & 86.3 & 60.1 & 69.4 \\
& Citrus-V       & 8B  & 85.5 & 95.5 & 90.2 & 78.7 & 100.0 & 88.1 & 79.5 & 97.5 & 87.6 & 58.4 & 90.4 & 70.9 & 84.2 \\
& Lingshu        & 7B  & 83.1 & 94.3 & 88.4 & 91.4 & 94.1 & 92.8 & 86.0 & 93.3 & 89.5 & 84.0 & 84.4 & 84.2 & 88.7 \\
& \textbf{\method{}} & 7B  & 91.1 & 92.9 & \textbf{92.0} & 95.1 & 96.0 & \textbf{95.5} & 90.8 & 89.5 & \textbf{90.2} & 96.6 & 81.6 & \textbf{88.5} & \textbf{91.6} \\
\bottomrule
\end{tabular}
\end{table*}
To validate the effectiveness of \method{}, we conduct a series of experiments corresponding to each stage of the proposed framework introduced in Sec.~\ref{sec:3.methodology}. Each stage is validated on a task specifically aligned with its objective.
Stage~1 evaluates the discriminative capability of the learned \anotok{} tokens via \textbf{Medical Zero-shot Anomaly Detection}.
Stage~2 assesses temporal reasoning through the \textbf{MMXU Benchmark} for medical symptom tracking.
Finally, Stage~3 examines \textbf{Medical Visual Explainability}, evaluating region-level grounding and the consistency between visual evidence and textual reasoning using segmentation-based metrics.

As this section focuses on the stage-specific tasks central to our framework, evaluations on conventional medical VQA benchmarks (\textbf{VQA RAD}~\cite{vqa_rad}, \textbf{SLAKE}~\cite{slake}, \textbf{PathVQA}~\cite{pathvqa}, and \textbf{MMMU Med}~\cite{mmmu}) are provided  in Appendix Sec.~\ref{sec:C.exp}, demonstrating that \method{} successfully preserves general medical knowledge of backbone while simultaneously reinforcing its stage-wise clinical applicability.

\subsection{Medical Zero-shot Anomaly Detection}
\label{subsec:4.1.anomaly}
\newcolumntype{O}{>{\columncolor{gray!10}}c}

\renewcommand{\arraystretch}{0.85}
\begin{table*}[t]
\small
\centering
\caption{Results on MMXU~\cite{mmxu}. Models are categorized into general-purpose, closed-source, and medical-domain VLMs.}
\setlength{\tabcolsep}{1.0em}{
\begin{tabular}{l|l|c|ccc|O}
\toprule
\textbf{Category} & \textbf{Model} & \textbf{Size} & \textbf{Worsen} & \textbf{Improved} & \textbf{No Change} & \textbf{Overall}~($\uparrow$) \\
\midrule

\multirow{2}{*}{\textbf{General}} 
& InternVL2.5 & 8B & 0.486 & 0.607 & 0.402 & 0.498 \\
& Qwen2.5-VL   & 7B & 0.499 & 0.545 & 0.513 & 0.519 \\
\midrule

\multirow{2}{*}{\textbf{Closed}} 
& Claude-3.5 & -- & 0.494 & 0.518 & 0.493 & 0.502 \\
& GPT-4o     & -- & 0.480 & 0.675 & 0.559 & 0.571 \\
\midrule

\multirow{3}{*}{\textbf{Medical}} 
& Citrus-V   & 8B & 0.468 & 0.713 & 0.532 & 0.571 \\
& Lingshu    & 7B & 0.597 & \textbf{0.734} & 0.529 & 0.620 \\
& \textbf{\method{} (Ours)} & 7B & \textbf{0.663} & 0.714 & \textbf{0.588} & \textbf{0.655} \\
\bottomrule
\end{tabular}}
\label{tab:mmxu}
\end{table*}

\paragraph{Experimental Settings.}
In the zero-shot anomaly detection setting, the model is tested on datasets that are entirely unseen during training to ensure a fair comparison with competing models. 
We evaluate across four heterogeneous modalities—\textbf{Brain MRI}, \textbf{Head CT}, \textbf{Br35h}, and \textbf{COVID-19 X-ray}~\cite{brainmri, headct, Br35h, covid19}—to assess generalization capability. 
Each test sample is queried with a consistent instruction prompt:
\textit{``Is there any abnormality in this image?''} 
The model’s binary response is evaluated using the F1 score to capture both precision and recall performance.

We compare \method{} against a range of baselines spanning three categories: (1) \textbf{General-purpose open-source VLMs:} Qwen2.5-VL-7B and InternVL2.5-8B~\cite{qwen2.5vl, internvl2.5},  
(2) \textbf{Closed-source models:} GPT-4o and Claude-3.5~\cite{gpt4o, claude},  
and (3) \textbf{Anomaly Detection Specialized VLMs:} AnomalyGPT and Anomaly-OV~\cite{anomaly_gpt, anomalyov}, as well as (4) \textbf{Medical Foundation VLMs:} LLaVA-Med-7B, Citrus-V-8B, and Lingshu-7B~\cite{llava_med, citrus_v, lingshu}.
\vspace{-1.0em}
\paragraph{Results.}
As summarized in Tab.~\ref{tab:tab1}, \method{} achieves \textbf{SOTA performance} across all four datasets, demonstrating superior generalization to unseen medical imaging modalities and conditions. 
The results indicate that the learned anomaly-aware tokens, \anotok{}, effectively capture lesion-relevant features, enabling reliable and robust zero-shot abnormality discrimination even in unseen datasets. 
Notably, \method{} surpasses all medical-specialized baselines and even outperforms closed-source models, validating the efficacy of its anomaly representation learning. 
While some closed-source models (e.g., GPT-4o) occasionally produce conservative outputs for normal cases due to safety-alignment bias~\cite{xie2024sorry,mmpb} (e.g., responding with \textit{“I'm unable to analyze medical images ..."}), \method{} consistently produces well-calibrated predictions across both normal and abnormal cases, indicating greater clinical reliability and decision consistency. Overall, Stage~1 demonstrates that anomaly representation learning offers a reliable foundation for zero-shot medical abnormality detection.

\subsection{Medical Symptom Tracking}
\label{subsec:4.2.mmxu}
\paragraph{Experimental Settings.}
We evaluate temporal reasoning ability using the \textbf{MMXU} benchmark~\cite{mmxu}, which involves paired chest X-ray studies from the same patient.
For each instance, the model must classify disease progression as \textit{worsened}, \textit{improved}, or \textit{unchanged} based on two images and a multiple-choice question such as:\textit{“What is the condition of the left lower lung zone across the two chest CXR images? A: No significant change, B: Improved, C: Worsened.”}
Only models supporting multi-image reasoning are included in this comparison, and the evaluation metric follows the accuracy of categorical predictions derived from the model’s textual responses. 
\renewcommand{\arraystretch}{0.85}
\begin{table*}[t]
\small
\centering
\caption{Visual grounding performance on BMAD~\cite{bmad} (BraTS2021~\cite{brats2021}, RESC~\cite{resc}, BTCV + LiTs~\cite{btcv, lits}) and ChestX-Det~\cite{chestxdet} datasets. Each dataset reports AUC and mIoU metrics (higher is better).}
\setlength{\tabcolsep}{0.45em}{
\begin{tabular}{lcccccccc}
\toprule
\multirow{2}{*}{\textbf{Model}} &
\multicolumn{2}{c}{\textbf{BraTS2021}} &
\multicolumn{2}{c}{\textbf{RESC}} &
\multicolumn{2}{c}{\textbf{BTCV + LiTs}} &
\multicolumn{2}{c}{\textbf{ChestX-Det}} \\
\cmidrule(lr){2-3} \cmidrule(lr){4-5} \cmidrule(lr){6-7} \cmidrule(lr){8-9}
 & \textbf{AUC} ($\uparrow$) & \textbf{mIoU} ($\uparrow$)
 & \textbf{AUC} ($\uparrow$) & \textbf{mIoU} ($\uparrow$)
 & \textbf{AUC} ($\uparrow$) & \textbf{mIoU} ($\uparrow$)
 & \textbf{AUC} ($\uparrow$) & \textbf{mIoU} ($\uparrow$) \\
\midrule
Citrus-V                 & 98.8 & 32.6 & 87.6 & 2.1 & 82.1 & 1.7 & 98.0 & 12.4 \\
\textbf{Medic-AD (Ours)} & \textbf{99.8} & \textbf{87.6} & \textbf{100} & \textbf{97.2} & \textbf{97.2} & \textbf{83.6} & \textbf{99.8} & \textbf{79.8} \\
\bottomrule
\end{tabular}}
\label{tab3:vg}
\end{table*}

\vspace{-1.0em}
\paragraph{Results.}
As shown in Tab.~\ref{tab:mmxu}, \method{} demonstrates clear advantages over existing multimodal models on the MMXU benchmark. By leveraging the \difftok{} tokens representations disentangled through the Diff Q-Former, the model effectively captures localized lesion changes while being robust to irrelevant appearance variations such as illumination or positional shifts. Unlike general VLMs that simply concatenate multiple images, \method{} explicitly encodes inter-image relationships, yielding consistent reasoning about temporal dynamics.  In particular, qualitative inspection shows that \method{} highlights pathological regions with true clinical changes (e.g., consolidation growth or opacity reduction), whereas other models often mistake global contrast shifts for disease progression (Fig.\ref{fig2}). These findings indicate that Stage~2 effectively separates clinically relevant temporal changes.

\begin{figure}[t]
    \centering
    \begin{subfigure}[t]{0.48\linewidth}
        \centering
        \includegraphics[width=\linewidth]{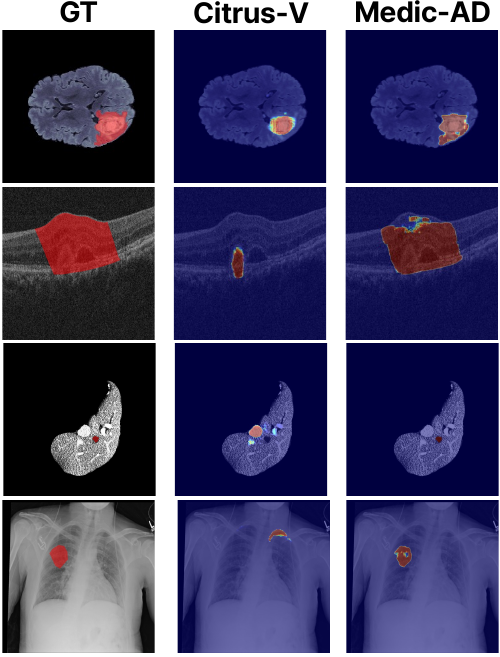}
        \caption{Abnormal samples}
        \label{fig:abnormal}
    \end{subfigure}
    \hspace{-0.002\linewidth}
    \vrule width 0.8pt 
    \hspace{0.007\linewidth}
    \begin{subfigure}[t]{0.48\linewidth}
        \centering
        \includegraphics[width=\linewidth]{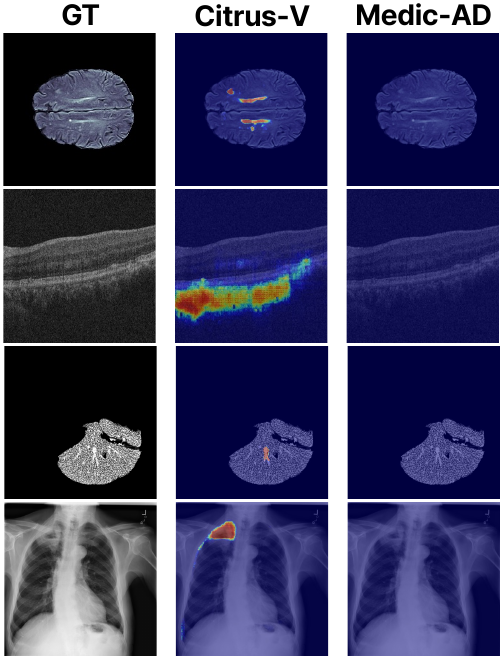}
        \caption{Normal samples}
        \label{fig:normal}
    \end{subfigure}
    \caption{Visual Grounding comparison between \method{} and Citrus-V~\cite{citrus_v} on diverse  abnormal and normal samples.}
    \label{fig4}
    \vspace{-1.5em}
\end{figure}

\subsection{Medical Visual Explainability}
\label{subsec:4.3.seg}
\paragraph{Experimental Settings.}
To assess visual grounding and explainability, we evaluate \method{} on medical datasets that include pixel-level anomaly masks, specifically a subset of \textbf{BMAD}~\cite{bmad} (BraTS2021~\cite{brats2021}, RESC~\cite{resc}, and BTCV + LiTs~\cite{btcv, lits}) and the \textbf{ChestX-Det}~\cite{chestxdet} dataset. 
For each image, the model generates a heatmap using the same anomaly-detection query as in Sec.~\ref{subsec:4.1.anomaly}.
Predicted heatmaps are compared with ground-truth masks using AUC and mIoU.
We compare against \textbf{Citrus-V}~\cite{citrus_v}, the most recent medical VLMs supporting visual grounding.
\vspace{-1.0em}
\paragraph{Results.}
\method{} shows consistently strong performance across datasets, outperforming Citrus-V on both AUC and mIoU (Tab.~\ref{tab3:vg}).
By integrating \anotok{} tokens with intermediate visual features, \method{} produces heatmaps that more accurately localize pathological regions aligned with the model’s textual rationale.
In contrast, Citrus-V, which use a SAM2 decoder~\cite{sam2}, tends to produce less precise masks, sometimes highlighting non-lesion areas or yielding diffuse activations in normal images.
Representative examples in Fig.~\ref{fig4} highlight these differences. These results show that Stage~3 improves the alignment between visual evidence and model reasoning, yielding more clinically coherent responses.

\section{Analysis}
\label{sec:5.Analysis}
In this section, we present a comprehensive analysis of the proposed \method{} framework through ablation studies, hyperparameter sensitivity experiments, and real-world evaluations. Across all studies, the results consistently support the central claim of this work: temporal reasoning in medical image pairs fundamentally benefits from coherent integration of anomaly-aware spatial cues and temporally grounded difference representations.

\subsection{Effect of \textbf{\anotok{}} Tokens on Temporal Reasoning}
\label{sec:5.1.Analysis}
The first analysis focuses on how \anotok{} tokens contributes to constructing reliable temporal representations.
In \method{}, the construction of \difftok{} tokens is grounded in the anomaly-aware visual features generated during the \anotok{} tokens estimation process, where patch-wise anomaly likelihood modulates the magnitude of visual representations (Sec.~\ref{subsec:3.2.training_pipelines}). To isolate the role of \anotok{} tokens, we generate \difftok{} tokens using the unmodified, original visual features without salience adjustment.

As reported in Tab.~\ref{tab:ano_visual_feature} (\textit{Effect of \anotok{} Tokens}), removing \anotok{} tokens consistently degrades performance across tasks. The drop reveals two important observations. First, anomaly-aware magnitude modulation enhances the expressiveness of the visual embeddings used for temporal differencing, allowing \difftok{} tokens to capture clinically relevant cues rather than global appearance changes. Second, incorporating \anotok{} tokens during inference adds complementary contextual information that stabilizes the interpretation of new findings. Together, these results show that \anotok{} and \difftok{} tokens form a mutually reinforcing pair: one grounds spatial anomaly cues, while the other captures temporal evolution, and both are required for accurate modeling of medical image progression.
\renewcommand{\arraystretch}{0.85}
\begin{table}[t]
\small
\centering
\caption{Effect of utilizing \anotok{} tokens and comparison of visual feature extraction strategies.}
\vspace{-0.5em}
\setlength{\tabcolsep}{0.55em}{
\begin{tabular}{lccc}
\toprule
\text{} & \textbf{\anotok{}} & \textbf{Avg. F1~($\uparrow$)} & \textbf{MMXU~($\uparrow$)} \\
\midrule[0.75pt]
\multicolumn{4}{l}{\textit{Effect of \anotok{} Tokens}} \\
\midrule
\difftok{} tokens only        & $\times$ & --    & 0.635 \\
\method{}              & $\checkmark$ & --    & \textbf{0.655} \\
\midrule[0.75pt]
\multicolumn{4}{l}{\textit{Feature Selection Strategy}} \\
\midrule
Last layer             & $\checkmark$ & 90.9  & 0.619 \\
Intermediate 4-layers  & $\checkmark$ & \textbf{91.6} & \textbf{0.635} \\
\bottomrule
\end{tabular}}
\label{tab:ano_visual_feature}
\end{table}

\begin{figure}[t]
    \centering
    \begin{subfigure}[t]{0.495\linewidth}
        \centering
        \includegraphics[width=\linewidth]{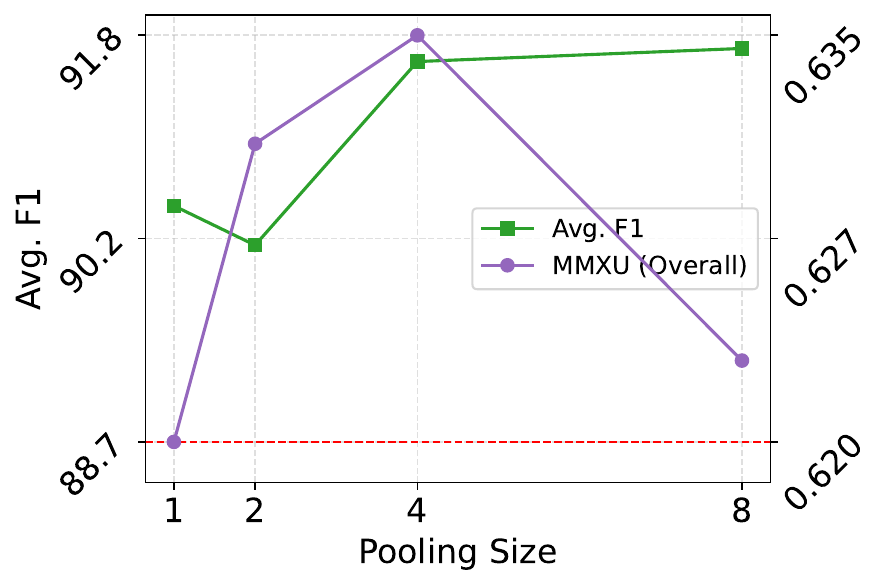}
        \caption{}
        \label{fig:pooling}
    \end{subfigure}
    \hspace*{-0.85em} 
    \begin{subfigure}[t]{0.495\linewidth}
        \centering
        \includegraphics[width=\linewidth]{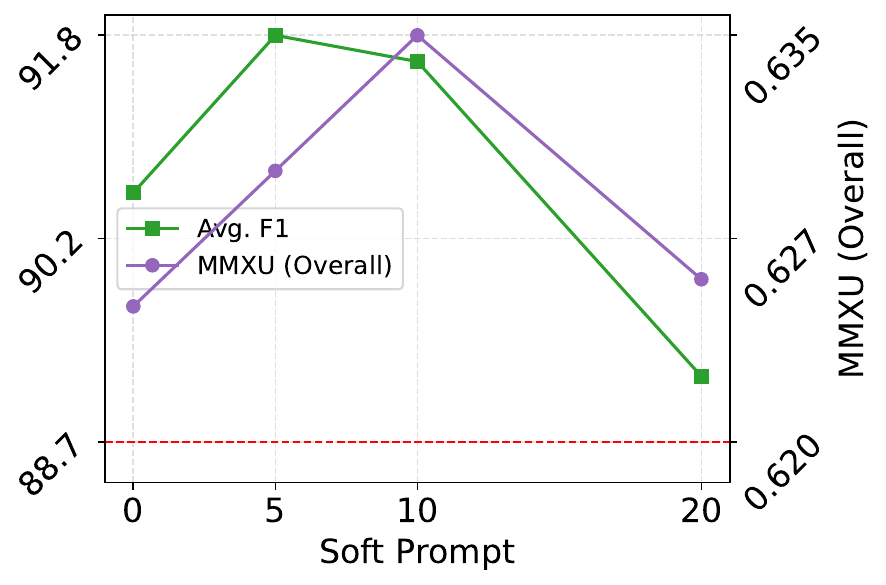}
        \caption{}
        \label{fig:soft_prompt}
    \end{subfigure}
    \vspace{-0.5em}
    \caption{Hyperparameter sensitivity analysis on (a) query token pooling size and (b) visual soft prompt counts. The red line denotes the baseline performance of Lingshu~\cite{lingshu}.}
    \label{fig5}
    \vspace{-1.0em}
\end{figure}

\subsection{Layer Selection in Visual Feature Extraction}
\label{sec:5.3.Analysis}
In generating both \anotok{} and \difftok{} tokens, \method{} relies on visual features extracted from the vision encoder of the backbone VLM. To understand how visual representations influence anomaly and temporal token construction,
we compare two configurations: using only the last-layer visual features, and aggregating intermediate-layer features together with the final representation.

As shown in Tab.~\ref{tab:ano_visual_feature}~(\textit{Feature Selection Strategy}), using intermediate-layer features consistently outperforms relying solely on the final hidden state. This advantage echoes prior findings~\cite{adaclip, anomalyclip, mvda, anomalyov} that multi-level feature aggregation provides a broader range of semantic cues and improves downstream performance. In our setting, incorporating intermediate features not only enriches the anomaly representations but also produces feature embeddings that are more stable and informative for computing inter-image differences in temporal reasoning.

\subsection{Impact of Hyperparameters}
\label{sec:5.3.Analysis}
\method{} involves two key hyperparameters: (1) the number of generated \anotok{} tokens and \difftok{} tokens, and (2) the number of soft prompts injected into the vision encoder. The number of \anotok{} and \difftok{} tokens is implicitly determined by the 2D pooling size applied to the magnitude-adjusted visual features used for token generation. Therefore, we investigate the effect of varying the pooling size. In parallel, the number of soft prompts influences how the extracted visual features are adapted to \method{} while also affecting the overall performance of the backbone VLM, motivating a sensitivity analysis.

As illustrated in Fig.~\ref{fig5}~(a), the model achieves consistently strong performance on both Anomaly Detection and MMXU benchmarks when the query-token pooling size is set to \(4 \times 4\), indicating that this granularity provides a favorable balance between spatial abstraction and anomaly localization. Similarly, the soft prompt sensitivity study in Fig.~\ref{fig5}~(b) demonstrates that using 10 visual soft prompts yields the most stable and competitive results. We therefore adopt a pooling size of \(4 \times 4\) and 10 soft prompts as the default configuration for \method{}.



\renewcommand{\arraystretch}{0.85}
\begin{table}[t]
\small
\centering
\caption{Evaluation on a real-world clinical dataset of 300 patients using GREEN~\cite{green}, RaTEScore~\cite{ratescore}, and GPT-4o evaluation.}
\vspace{-0.5em}
\setlength{\tabcolsep}{0.9em}{
\begin{tabular}{lccc}
\toprule
\textbf{Model} & \textbf{GREEN} & \textbf{RaTEScore} & \textbf{GPT eval} \\
\midrule
Lingshu-7B & 0.009 & 0.359 & 0.177 \\
\textbf{\method{}} 
& \textbf{0.020} & \textbf{0.430} & \textbf{0.291} \\
\bottomrule
\end{tabular}}
\label{tab5:smc}
\vspace{-1.0em}
\end{table}

\subsection{Real-world Application}
\label{sec:5.5.real}
To validate the applicability of \method{} beyond public benchmarks, we conduct a real-world clinical study using chest X-ray pairs collected from 300 patients who visited a hospital for follow-up examinations.
For each image pair, radiologists provide structured annotations describing the presence or absence of specific clinical findings, and the degree of change compared to the previous examination. Using this dataset, we formulate a temporal-difference captioning task in which the model must generate clinically consistent descriptions of symptom progression.

As shown in Tab.~\ref{tab5:smc}, \method{} outperforms Lingshu~\cite{lingshu}—the strongest baseline in temporal reasoning (Tab.~\ref{tab:mmxu})—under GREEN~\cite{green}, RaTEScore~\cite{ratescore}, and GPT evaluation. These results indicate that \method{} remains effective not only in controlled benchmarks but also demonstrates robustness and reliability on real-world clinical data. Moreover, the model produces descriptions that align closely with expert assessments, highlighting its potential for integration into clinical workflows that demand explainable and accurate temporal reasoning.
\section{Conclusion}
\label{sec:6.conclusion}
We introduced \textbf{\method{}}, a stage-wise medical VLM that strengthens clinical intelligence: lesion detection, temporal reasoning, and visual explainability through anomaly-aware and difference-token mechanisms. Our unified design enables the model to focus on abnormality cues, capture clinically meaningful changes between images, and provide accurate grounded visual evidence. Furthermore, evaluations on real-world hospital cases show robust alignment with expert assessments, indicating that \method{} offers a practical and reliable application for clinically usable medical VLMs.
\newpage
\section*{Acknowledgements}
\label{sec:7.Ack}
This work was supported in part by National Research Foundation of Korea (NRF) grant (RS-2024-00414981), Institute of Information \& communications Technology Planning \& Evaluation (IITP) grant (RS-2025-25442338, RS-2024-00397085, RS-2021-II211343), and by the Health and Medical R\&D Program of the Ministry of Health and Welfare (RS-2025-25455059). This research was supported by the AI Computing Infrastructure Enhancement (GPU Rental Support) User Support Program funded by the Ministry of Science and ICT (MSIT), Republic of Korea(RQT-25-090156). This research was also conducted as part of the Sovereign AI Foundation Model Project (Data Track, 2026-AIData-WII01), organized by the Ministry of Science and ICT (MSIT) and supported by the National Information Society Agency (NIA). We also thank the support from NVIDIA AI Technology Center (NVAITC), Samsung Changwon Hospital, and Samsung Medical Center. J. Do is with ASRI, Seoul National University.
{
    \small
    \bibliographystyle{ieeenat_fullname}
    \bibliography{main}
}

\clearpage
\appendix
\setcounter{page}{1}
\maketitlesupplementary

This supplementary document provides additional details supporting the main paper.

\section{Chat Template}
\label{sec:A.chat}

\begin{figure}[h]
    \centering
    \includegraphics[width=0.5\textwidth]{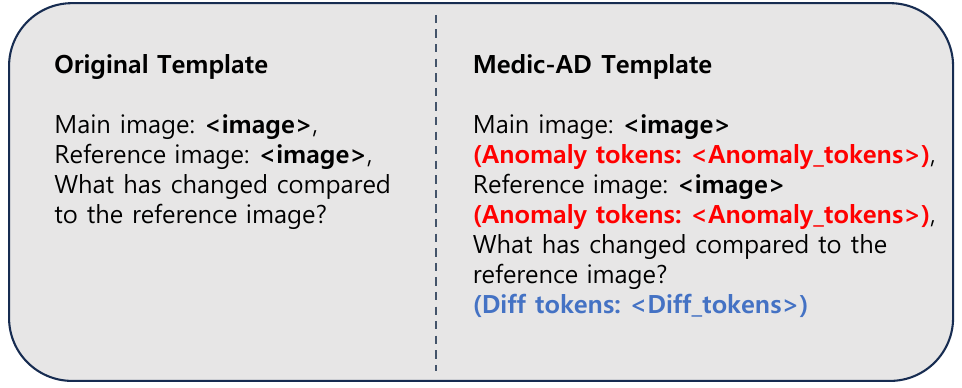}
    \caption{Comparison of chat template between original and \method{}, utilizing \anotok{} and \difftok{} tokens.}
    \label{fig_a_1}
\end{figure}

To incorporate the new \anotok{} and \difftok{} tokens, we update the original chat template to provide more explicit cues for abnormality detection and change description (Fig.~\ref{fig_a_1}). While the original template simply presented two images and asked what had changed, \method{} template attaches anomaly tokens (\anotok{}) to each image and adds diff tokens (\difftok{}) to the change question. This enriches the prompt with clinically aligned signals, enabling the model to produce more accurate abnormality identification and localized change explanations.

\section{Training Details}
\label{sec:B.training}
All training stages employ the AdamW optimizer together with a cosine learning rate scheduler.
\textbf{Stage~1} is trained on 3~H200 GPUs for one epoch (3~hours) using an initial learning rate of $1e-6$ and a per-device batch size of 16, resulting in a global effective batch size of 48.
\textbf{Stage~2} is trained on the same hardware for one epoch (6~hours) using an initial learning rate of $1e-6$ and a global effective batch size of 96.
Both Stage~1 and Stage~2 are optimized using the standard LLM cross-entropy loss.
\textbf{Stage~3} is trained for 100~epochs (15~hours) using an initial learning rate of $1e-3$ and a global effective batch size of 96, optimizing the DiceCE loss.
The parameter size of additional modules introduced in \method{} follows the specifications summarized in Tab.~\ref{tab:B_config}.
\begin{table}[h]
\centering
\small
\caption{Parameter size of \method{}'s additional modules.}
\label{tab:B_config}
\setlength{\tabcolsep}{6pt}
\renewcommand{\arraystretch}{0.95}
\begin{tabular}{lc}
\toprule
\textbf{Category} & 
\textbf{Param. Size} \\
\midrule
Visual soft prompts & 51K \\
\midrule
\anotok{} modules & 40M \\
\midrule
\difftok{} modules & 40M \\
\midrule
Heatmap modules & 48M \\
\bottomrule
\end{tabular}
\end{table}

\section{Additional experiments}
\label{sec:C.exp}

\subsection{Generic-Medical VQA}
In this section, we assess the general medical reasoning capability of \method{} by benchmarking it on a diverse set of established medical VQA and QA datasets, including VQA-RAD~\cite{vqa_rad}, SLAKE~\cite{slake}, PathVQA~\cite{pathvqa}, MMMU~Med~\cite{mmmu}, PMC-VQA~\cite{pmcvqa} for visual reasoning, and MedMCQA~\cite{medmcqa}, PubMedQA~\cite{pubmedqa}, MedQA~\cite{medqa}, MedXpertQA~\cite{medxpertqa} for text-only medical QA. As shown in Tab.~\ref{tab:C_vqa} and Tab.~\ref{tab:C_qa}, \method{} preserves broad medical knowledge acquired during pretraining while maintaining strong question--answering competence across both visual and non-visual domains. Notably, when comparing against a fully fine-tuned Lingshu baseline—trained end-to-end on the Stage~1 and Stage~2 datasets—we observe a pronounced degradation in general medical knowledge, leading to substantial drops across benchmarks. This contrast underscores that \method{}'s stage-wise design, which updates only targeted components such as \anotok{} and \difftok{}, effectively avoids catastrophic forgetting and retains general-domain medical reasoning capabilities.

\subsection{Results on hyperparameter experiments}
The experimental results on hyperparameters summarized in Fig.~\ref{fig5} are provided in full detail in Tab.~\ref{tab:C_ablation}, offering a comprehensive breakdown of each experimental configuration.

\subsection{Stage-wise vs.\ Joint Training}
Tab.~\ref{tabR2:stagewise_joint} shows that stage-wise training consistently outperforms joint optimization under identical experimental settings. 
While joint training still yields performance improvements over the baseline, stage-wise optimization achieves superior results by providing a more stable learning curriculum. 
This result suggests that the proposed curriculum enables each stage-specific objective to be effectively internalized before introducing additional supervision signals. 

\renewcommand{\arraystretch}{0.85}
\begin{table*}[h]
\small
\centering
\caption{Results on medical VQAs (VQA RAD~\cite{vqa_rad}, SLAKE~\cite{slake}, PathVQA~\cite{pathvqa}, MMMU Med~\cite{mmmu}), and PMC VQA~\cite{pmcvqa}).}
\setlength{\tabcolsep}{1.0em}{
\begin{tabular}{l|l|cccccc}
\toprule
\textbf{Model} & \textbf{Size} & \textbf{VQA RAD} & \textbf{SLAKE} & \textbf{PathVQA} & \textbf{MMMU Med} & \textbf{PMC VQA} \\
\midrule
LLaVA-Med  & 7B & 53.7 & 48.0 & 32.5 & 29.3 & 30.5 \\
Citrus-V   & 8B & 64.3 & \textbf{84.9} & \textbf{62.0} & 46.4 & 55.6 \\
Lingshu    & 7B & \textbf{67.9} & 83.1 & 61.9 & 54.0 & 56.0 \\
\midrule
Lingshu (full finetuning)    & 7B & 40.8 & 38.4 & 45.3 & 24.8 & 6.4 \\
\textbf{\method{}} & 7B & 64.3 & 78.5 & 56.5  & \textbf{54.2} & \textbf{56.1} \\
\bottomrule
\end{tabular}}
\label{tab:C_vqa}
\end{table*}
\renewcommand{\arraystretch}{0.85}
\begin{table*}[h]
\small
\centering
\caption{Results on medical QAs (MedMCQA~\cite{medmcqa}, PubMedQA~\cite{pubmedqa}, MedQA~\cite{medqa}, and MedXpertQA~\cite{medxpertqa}).}
\setlength{\tabcolsep}{1.0em}{
\begin{tabular}{l|l|cccccc}
\toprule
\textbf{Model} & \textbf{Size} & \textbf{MedMCQA} & \textbf{PubMedQA} & \textbf{MedQA} & \textbf{MedXpertQA}\\
\midrule
LLaVA-Med  & 7B & 39.4 & 26.4 & 42.0 & 9.9\\
Citrus-V   & 8B & 55.1 & 74.8 & \textbf{64.9} & \textbf{16.9} \\
Lingshu    & 7B & 55.9 & 75.4 & 63.3 & 16.5 \\
\midrule
Lingshu (full finetuning)    & 7B & 1.8 & 35.4 & 24.4 & 10.9 \\
\textbf{\method{}} & 7B &\textbf{56.7} & \textbf{75.6} & 63.6 & 16.5 \\
\bottomrule
\end{tabular}}
\label{tab:C_qa}
\end{table*}
\begin{table*}[h]
\centering
\small
\caption{Ablation on pooling size and soft prompt counts. We report average F1 and MMXU (Overall) for each configuration.}
\label{tab:C_ablation}
\setlength{\tabcolsep}{6pt}
\renewcommand{\arraystretch}{0.95}
\begin{tabular}{lccc}
\toprule
\textbf{Category} & \textbf{Config. Value} & \textbf{Avg. F1} & \textbf{MMXU} \\
\midrule
\multirow{4}{*}{\textbf{Pooling Size}} 
& 1 & 90.5 & 0.620 \\
& 2 & 90.2 & 0.631 \\
& 4 & 91.6 & 0.635 \\
& 8 & 91.7 & 0.623 \\
\midrule
\multirow{4}{*}{\textbf{Soft Prompt Counts}} 
& 0  & 90.6 & 0.625 \\
& 5  & 91.8 & 0.630 \\
& 10 & 91.6 & 0.635 \\
& 20 & 89.2 & 0.626 \\
\bottomrule
\end{tabular}
\end{table*}

\renewcommand{\arraystretch}{0.85}
\begin{table*}[h]
\small
\centering
\caption{Comparison between Stage-wise and Joint training.}
\vspace{-0.5em}
\setlength{\tabcolsep}{0.8em}{
\begin{tabular}{lcc}
\toprule
\textbf{Model} & \textbf{AVG. F1} & \textbf{MMXU} \\
\midrule
\text{Lingshu-7B}       & 88.7 & 0.620 \\
\text{Medic-AD (Joint)}       & \underline{89.7} & \underline{0.630} \\
\textbf{Medic-AD (Stage-wise)} & \textbf{91.6} & \textbf{0.655} \\
\bottomrule
\end{tabular}}
\label{tabR2:stagewise_joint}
\vspace{-1.0em}
\end{table*}

\end{document}